\documentclass[runningheads]{llncs}
\usepackage{graphicx}
\usepackage{gensymb}
\usepackage{tikz}
\def\checkmark{\tikz\fill[scale=0.4](0,.35) -- (.25,0) -- (1,.7) -- (.25,.15) -- cycle;}

\usepackage{multirow}
\usepackage{url}
\usepackage{subfigure}
\usepackage{float}
\usepackage{textcomp}
\usepackage{gensymb}

\begin{document}
\title{UGformer for Robust Left Atrium and Scar Segmentation Across Scanners  
\thanks{This research is funded by XJTLU Research Development Funding 20-02-60. Computational resources used in this research are provided by the School of Robotics, XJTLU Entrepreneur College (Taicang), Xi'an Jiaotong-Liverpool University.}
}
% \author{Tianyi Liu\inst{1}\orcidID{0000-0003-2778-635X} \and
% Size Hou\inst{3}\orcidID{0000-0002-3408-2791} \and
% Jiayuan Zhu\inst{2}\orcidID{0000-0002-8031-0767} \and
% Zilong Zhao\inst{1}\orcidID{0000-0001-9262-4467} \and
% Haochuan Jiang*\inst{1}\orcidID{0000-0002-8727-4121}}
\author{Tianyi Liu\inst{1} \and
Size Hou\inst{3} \and
Jiayuan Zhu\inst{2} \and
Zilong Zhao\inst{1} \and
Haochuan Jiang*\inst{1}}

% %\author{First Author\inst{1}\orcidID{0000-1111-2222-3333} \and
% %Second Author\inst{2,3}\orcidID{1111-2222-3333-4444} \and
% %Third Author\inst{3}\orcidID{2222--3333-4444-5555}}
% %
\authorrunning{T. et al.}
% %\authorrunning{F. Author et al.}

\institute{School of Robotics \and 
School of Artificial Intelligence and Advanced Computing \\
XJTLU Entrepreneur College (Taicang), Suzhou, Jiangsu, 215412, P.R. China
\and
School of Science, Xi'an Jiaotong-Liverpool University, SIP, Suzhou, Jiangsu, 215123, P.R.China }
\maketitle              % typeset the header of the contribution
\begin{abstract}
% The abstract should briefly summarize the contents of the paper in
% 15--250 words.
Thanks to the capacity for long-range dependencies and robustness to irregular shapes, vision transformers and deformable convolutions are emerging as powerful vision techniques of segmentation.
Meanwhile, Graph Convolution Networks (GCN) optimize local features based on global topological relationship modeling. 
Particularly, they have been proved to be effective in addressing issues in medical imaging segmentation tasks including multi-domain generalization for low-quality images.
In this paper, we present a novel, effective, and robust framework for medical image segmentation, namely,  UGformer. 
It unifies novel transformer blocks, GCN bridges, and convolution decoders originating from U-Net to predict left atriums (LAs) and LA scars.
We have identified two appealing findings of the proposed UGformer: 1). an enhanced transformer module with deformable convolutions to improve the blending of the transformer information with convolutional information and help predict irregular LAs and scar shapes.
2). Using a bridge incorporating GCN to further overcome the difficulty of capturing condition inconsistency across different Magnetic Resonance Images scanners with various inconsistent domain information. 
The proposed UGformer model exhibits outstanding ability to segment the left atrium and scar on the LAScarQS 2022 dataset, outperforming several recent state-of-the-arts.
% \footnote{The code of the proposed UGformer will be released upon acceptance.}

\keywords{Left atrium segmentation, scar prediction, Transformer, Graph convolution model}
% \keywords{First keyword  \and Second keyword \and Another keyword.}
\end{abstract}
\section{Introduction}
Late gadolinium enhancement magnetic resonance imaging (LGE-MRI) is typically used to provide quantitative information on atrial scars~\cite{vergara2011tailored}. In this measurement, location and size in the left atrium (LA) indicate pathology (i.e., LA scars) and progression of atrial fibrillation~\cite{fudan.edu.cn}. 
 
\begin{figure}[!htb]
	\centering
	\subfigbottomskip = 2pt
	\subfigcapskip = 5pt
	\subfigure(a){
		\includegraphics[width=0.15\textwidth]{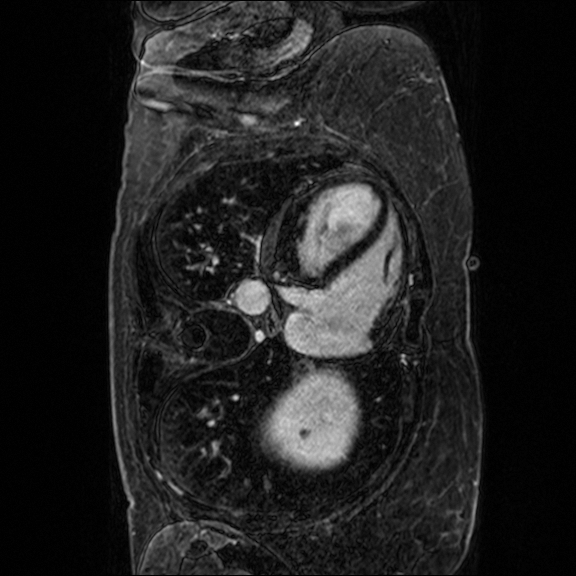}
		\label{SubFig:GoodContrast} }
	\subfigure(b){
		\includegraphics[width=0.15\textwidth]{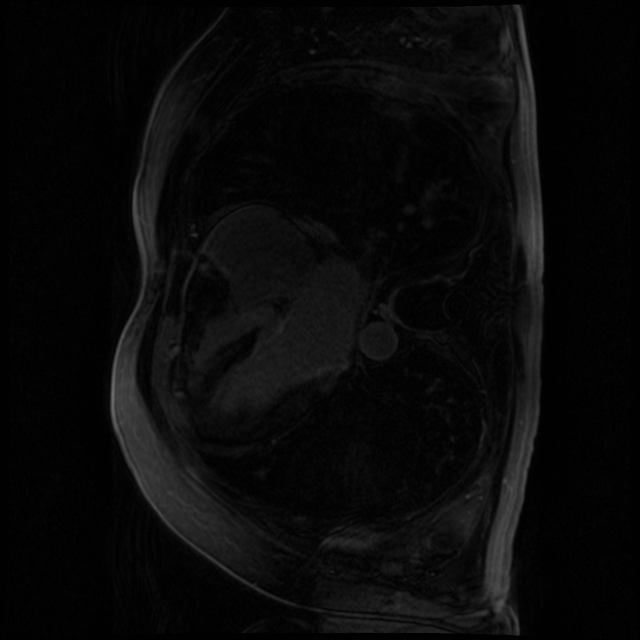}
		\label{SubFig:BadContrast} }  
	\subfigure(c){
		\includegraphics[width=0.15\textwidth]{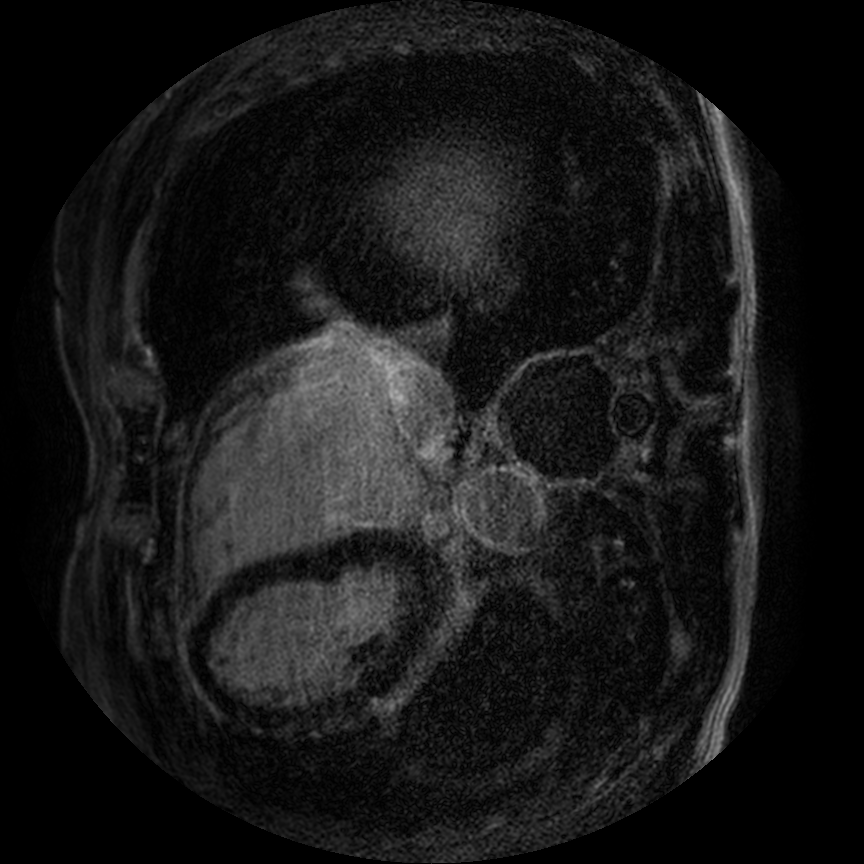}
		\label{SubFig:GoodRes} }
	\subfigure(d){
		\includegraphics[width=0.15\textwidth]{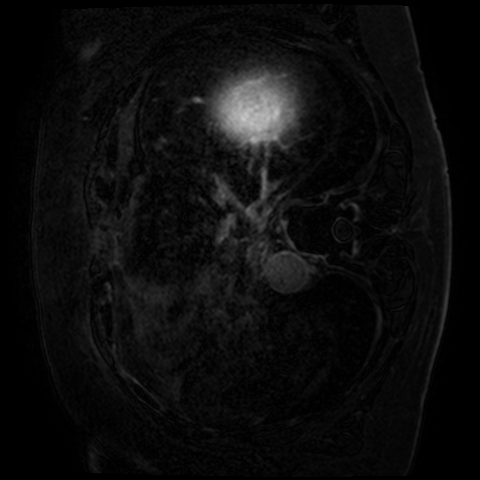}
		\label{SubFig:BadRes} }  
	\caption
	{Typical examples of LAScarQS Dataset~\cite{li2021atrialgeneral,li2022atrialjsqnet,li2022medical} in various contrast: (a) Proper contrast, (b) low contrast, and different spatial resolution (c) $886\times864$, (d) $480\times 480$.}
	\label{fig:DataExamples}
\end{figure}

Nowadays, deep learning models have been widely used to segment LA cavities and quantify LA scars from LGE-MRIs~\cite{chen2018multi} to help radiologists with initial screening for quick pathology detection. 
% To effectively leverage these data is crucial to segment LA and scars with deep learning models.
Meanwhile, LGE-MRIs are often collected by multiple scanners and possibly in low imaging quality. Each of them produces inconsistent domain information~\cite{li2021atrialgeneral}, including different contrast and spatial resolutions. (\ref{fig:DataExamples})
Promoting the generalization of a segmentation model against domain inconsistency becomes another challenge.

Essentially, semantic segmentation is a mapping from input images to output pixel labels through an empirically designed segmentation model. 
Recent computer vision research communities have witnessed great achievements brought by the Convolutional Neural Network (CNN) and Vision Transformers (ViT)~\cite{chen2021transunet,huang2021missformer}. However, there is a lack of theoretical explanations to guarantee prediction and generalization performance~\cite{JMLR:v23:21-0631}.
Besides, there is no fixed shape in human anatomies (i.e., LAs) and pathologies (i.e., LA scars). Atlas-based segmentation strategy cannot be utilized ideally~\cite{zhu2013automatic,li2020atrial}, while normal CNNs are not good at predicting deformable objects either~\cite{ronneberger2015u}.

Conventional CNN-based segmentation models only take care of local dependencies since the convolutional kernel only sees visual information in closing pixels within the receptive field. 
It leads to ignoring the full picture as a whole~\cite{raghu2021vision}. 
Common pooling layers in CNN will also degrade spatial information since it regards neighboring pixels as one single pixel. 
Losses in spatial information restrict the prediction performance of conventional CNN models~\cite{xiao2021early}.

Fortunately, Graph Convolutional Networks (GCN) are promised to address those challenges effectively by leveraging the robustness brought by the topological properties~\cite{kipf2016semi}. 
The topological relationship extracted by GCN while performing representation learning has been proved more stable against various application scenarios than that of the geometric relationship of general vision models, i.e., CNNs and ViTs~\cite{carlsson2020topological}. In addition to the local features extracted by CNNs, GCN also provides an approach to model the relationship among different local features. 
It optimizes local features of low-quality images by Laplacian smoothing to a certain extent~\cite{han2022vision}, beneficial to promoting generation across data from different domains. 

Meanwhile, recently ViT models are becoming popular in semantic segmentations in handling long-range dependencies. 
It models spatial image information by engaging the self-attention mechanism~\cite{vaswani2017attentionnet}. 
Swin Transformer~\cite{liu2021swin} and SegFormer~\cite{xie2021segformer} are two pioneering approaches to engaging ViTs in segmentation tasks. 
Swin Transformer engages sliding window operation. It fulfills the localization of convolutional operations while saving time consumption in computation. 
SegFormer connects the transformer to lightweight multi-layer perception decoders, allowing it to combine local and global attention.
In medical image segmentations, TransUnet~\cite{chen2021transunet}, UTnet~\cite{gao2021utnet}, and LeViT-Unet~\cite{xu2021levit} are the first few trials to integrate ViT modules in the U-Net~\cite{ronneberger2015u} architecture. All of them achieve state-of-the-art segmentation performance on the Synapse dataset~\cite{synapse}.

\begin{figure}
\includegraphics[width=\textwidth]{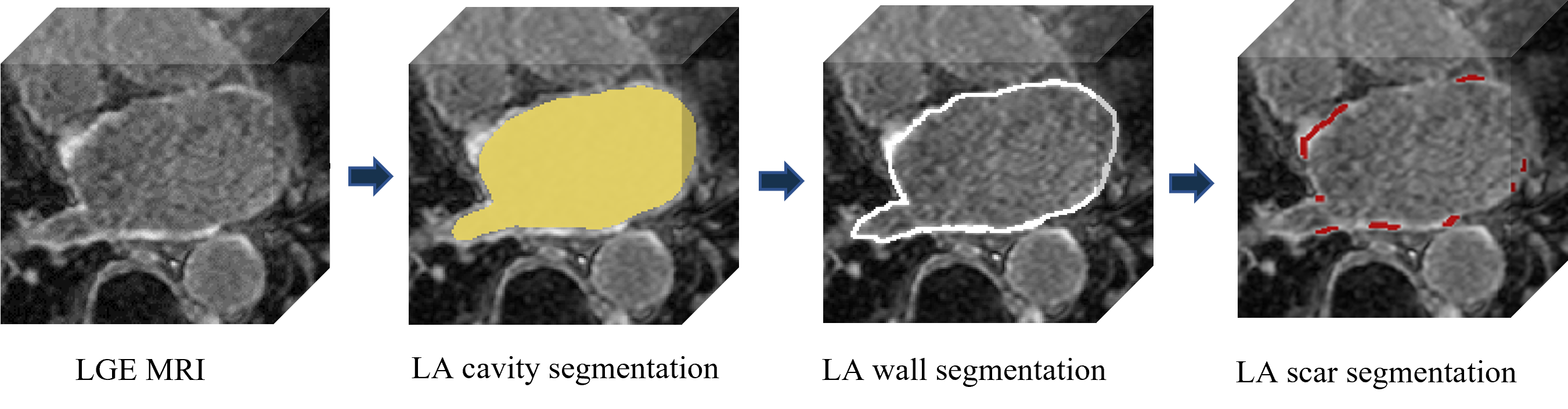}
\caption{Positions of LA and LA scars~\cite{li2022medical}} \label{fig:scar}
\end{figure}
In terms of LA scar prediction, prior work predicts LA and LA scars separately without considering the relationship between them~\cite{li2022medical}.
Meanwhile, the size of the scars is relatively insignificant, bringing difficulties in the prediction.
Fortunately, LAs are much easier to be predicted, while LA scars are often detected near identified LA boundaries Fig.~\ref{fig:scar}. 
Inspired by~\cite{zhou2021probabilistic}, we believe that combining the prediction of LAs and LA scars can be expected to improve scar segmentation performance.

In this paper, we propose a novel U-shaped GCN with Enhanced Transformer module (UGformer). It is a two-stage segmentation model by segmenting the LA before quantifying the irregularly shaped LA scars. 
It consists of a novel transformer block as the encoder, convolution blocks as the decoder, and skip-connections with a GCN as the bridge. 

In the encoder, the novel transformer block, namely, enhanced transformer block (ETB), is built by replacing the single multi-head self-attention module with paralleling the multi-head self-attention module (MHSA) and deformable convolutions (DCs). It models global spatial attention while dealing with irregular shape information by leveraging advantages in both convolutions and transformers, i.e., proper generalization ability and sufficient model capacity~\cite{xiao2021early}.
The bridge with GCN connection optimizes the fusion of long-range information and context information between the encoder and the decoder~\cite{han2022vision}. 
It continuously strengthens the representation of intermediate feature maps to find a low-dimensional invariant topology, improving the extrapolation of segmentation models. 

The major contributions of this paper are summarized as follows:
\begin{itemize}
  \item We proposed the UGformer, a novel two-stage segmentation model for LA and LA scar segmentation.
  \item In the encoder, we designed a novel enhanced transformer block combining multi-head self-attention and deformable convolutions to model global attention and address irregular shapes of LA scars.
  \item In the bridge, we proposed a novel GCN-based structure to optimize the global space of intermediate feature layers. 
  \item Compared to other state-of-the-art baselines, the predicting performance of the proposed model on LAScarQS dataset~\cite{li2021atrialgeneral,li2022atrialjsqnet,li2022medical} demonstrates the effectiveness and generalizability of the proposed UGformer.
\end{itemize}

\section{Methodology}
\begin{figure}[!htb]
\includegraphics[width=\textwidth]{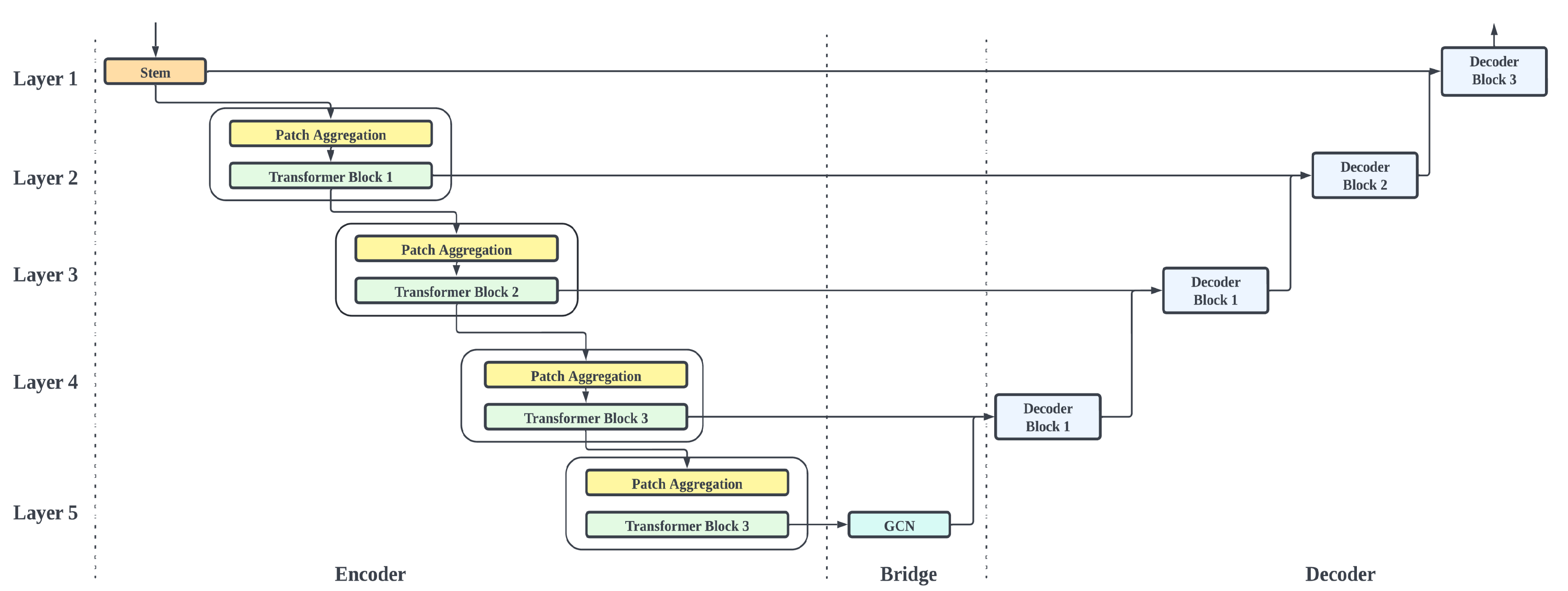}
\caption{UGformer Structure} \label{fig:model}
\end{figure}

As depicted in Fig.~\ref{fig:model}, the proposed UGformer consists of an encoder, a U-Net decoder~\cite{ronneberger2015u}, and a bridge. Specifically, the encoder is constructed by ETB, while deconvolutions are used to build the decoder. They are connected by the bridge with GCN.

\subsection{Encoder Block}
In the encoder, the convolutional STEM module~\cite{guo2022cmt}, including a convolution module, a GELU module, and a batchnorm to vectorize the input features with down-sampling, was employed. It promotes quick convergence and robustness during training.

Each encoding layer (seen in Fig.~\ref{fig:model}) is constructed by a Patch Aggregation Block. Be noted that the transformer operation is not designed to downsample the feature dimension. Instead, it is constructed by the Patch Aggregation Block, including a $2\times 2$ kernel and a stride operation with two steps to fulfill the hierarchy structure.

% but the model needs to construct downsampling to implement the hierarchy. To solve this problem, the Patch Aggregation Block, which contains a convolution kernel of 2x2 and a stride of 2, is used to achieve the effect of downsampling.

\begin{figure}[!htb]
\includegraphics[width=\textwidth]{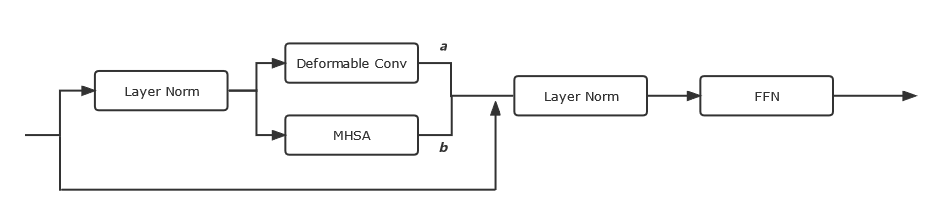}
\caption{EBT in UGformer} \label{fig:ETB}
\end{figure}

Besides, each layer also contains an ETB (seen in Fig.~\ref{fig:ETB}) to enable the UGformer to obtain both long-range dependencies and local context.

% A feed-forward module block is involved in the vanilla transformer~\cite{vaswani2017attention} to mix channel dimensional information.
% % The vanilla transformer consists of a LayerNorm, a multi-head self-attention (MHSA), another LayerNorm, and a feed forward module block to mix channel dimensional information.
% Although it allows convolutions to enhance local feature extraction. However, we argue that performing the channel dimensional feature transformation cannot make full use of local information. 
% Compared to~\cite{huang2021missformer} with $3\times3$ depthwise convolution with padding, the proposed ETB module employs interaction between the final convolutional layer with long-range dependencies and the layers before with local context.
% Therefore, we borrow the idea seen in~\cite{takikawa2019gated} to use a gate attention module (seen in Fig.~\ref{fig:gate})  to replace the feed-forward module to improve blending the transformer information with convolutional information.
Inspired from~\cite{vaswani2017attentionnet}, a single MHSA block is involved in ETB to extract long-range relationships and spatial dependencies. 
We engage DCs~\cite{dai2017deformable} parallel to MHSA to improve segmenting irregular LAs and quantifying LA scars.
% Having different shapes and angles of a certain object is a major challenge for detection/recognition. Deformable convolution adds an offset to each convolution sampling point, allowing the extraction of the complete irregular LA features. 
% Deformable convolution is here equivalent to an attention. 
To make ETB adapt to both MHSA and deformable convolutions, a set of learnable parameters ($a$ and $b$ see Fig.~\ref{fig:ETB}) are set to leverage both paralleling parts~\cite{pan2022integration}.

% Because of uncertainty as to which module is more suitable for extracting the attention of a particular layer(the MHSA module or the deformable module), we borrow the idea from ~\cite{pan2022integration} to set a set of learnable parameters to help the model perform batter. As 

% \begin{figure}
% \includegraphics[width=\textwidth]{gate.png}
% \caption{The gate module. $X$ denotes the feature map obtained from the patch aggregation module and $T$ represents the feature map obtained from the layernorm after multi-head self attention. $\oplus$ and $\otimes$ represent element-wise summation and mulitiplication respectively.
% } \label{fig:gate}
% \end{figure}

% $X$ is the feature map obtained from the patch aggregation module and $t$ is the feature map gotton from the layernorm after multi-head self attention. The gate attention can be formulated as:
% \begin{equation}
% F = Convset(concat(t,Linear(x)))
% \end{equation}
% where convset is a set of functions includes batchnorm, $1\times1$ convolution and relu in order.
% Then the feature map will do element-wise product with feature map from the transformer. To make sure the , it then do the residual adding. Finally the result will do the channel-wise weighting with kernel $w_{t}$.
% \begin{equation}
% F^{'}= Sigmoid(BN(Linear(F)))
% \end{equation}
% \begin{equation}
% T = ((F^{'} \odot x) + F^{'})^{T} \times w_{t} 
% \end{equation}

% \subsection{GCN Spectrum Clustering Bridge}
\subsection{Bridge}
%tianyi skip connection describe.
% In contrast to the previous idea of VITs outputting the structure directly through an MLP, 

\begin{figure}[!htb]\centering
\includegraphics[width=0.8\textwidth]{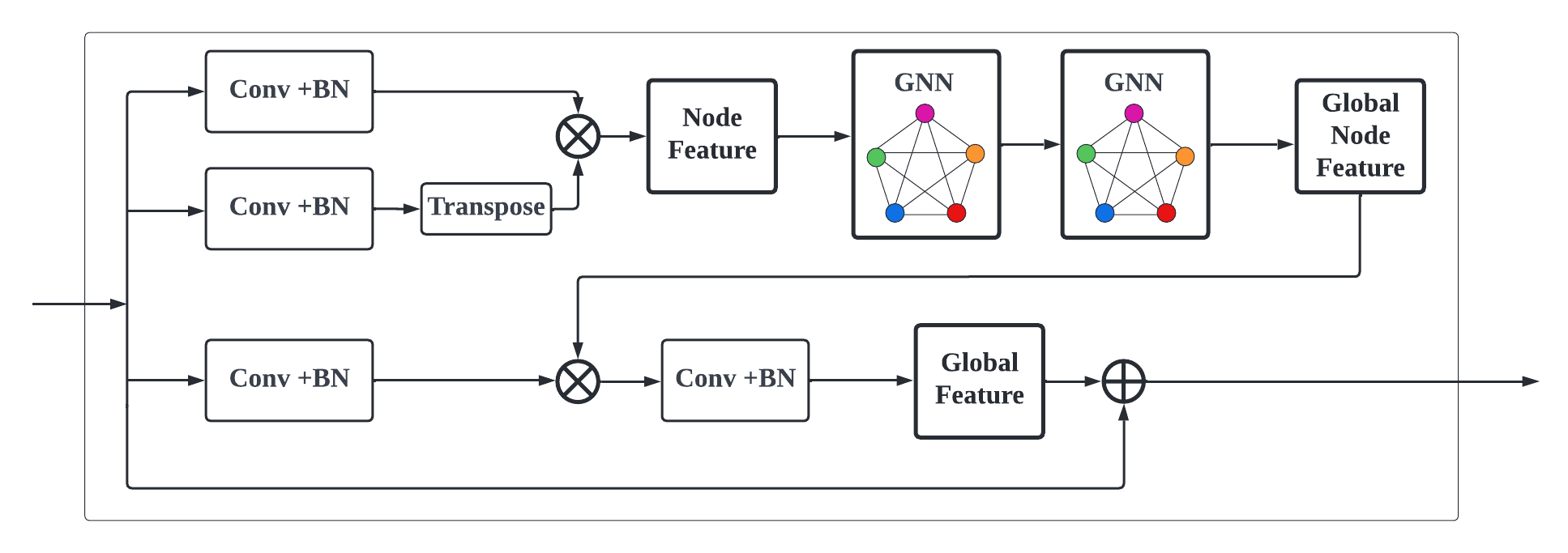}
\caption{The GCN Architecture in Fig.~\ref{fig:model}} \label{fig:GCNImplementation}
\end{figure}

The bridge module is added to the skip connection from the original U-Net~\cite{ronneberger2015u} with a GCN transformation (seen in Fig.~\ref{fig:GCNImplementation}). 
It bridges the encoder with ETB and the decoder constructed by convolutions to maximize the advantages brought by transformers and convolutions. 
It is capable of promoting the optimization of local features and generalization across data from different domains.
\begin{figure}[!htb]
	\centering
	\includegraphics[width=0.5\textwidth]{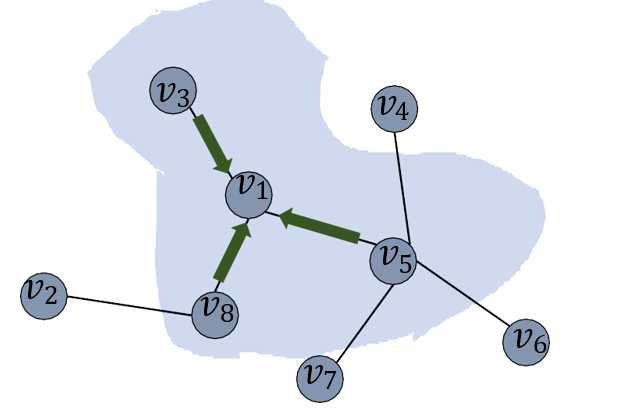}
	\caption
	{GCN Topology: the global relationship of graph-based feature structure. The arrows represent the closer relationship by GCN operations in the graph. The shadow represents the topology composed of the neighbors of node v1.}
	\label{fig:GCNDetails}
\end{figure}

\begin{figure}[!htb]
	\centering

	\includegraphics[width=1\textwidth]{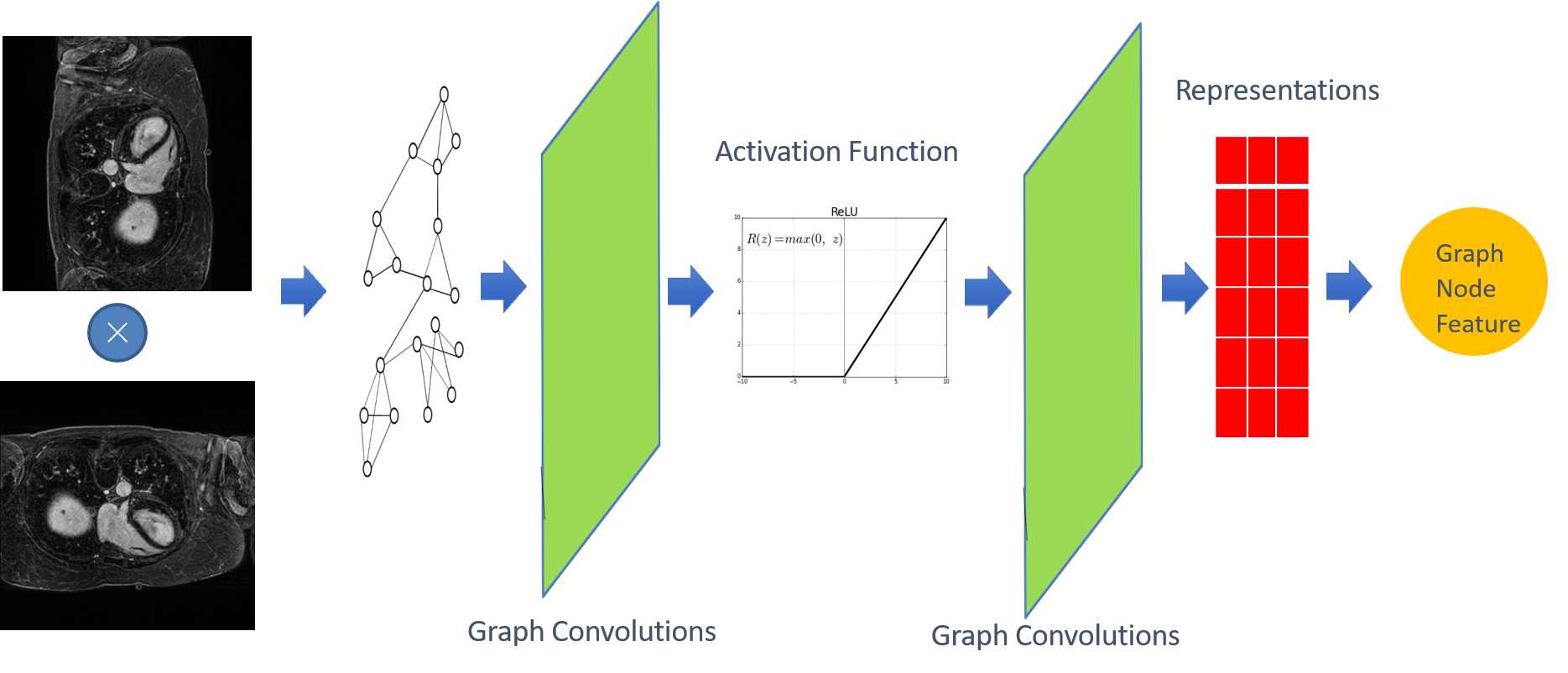}
	\caption
	{Two Layers of GCN Blocks: Input feature map multiplies its transpose and update by aggregation rules in GCN block~\cite{kipf2016semi}.}
	\label{fig:GCNDetails1}
\end{figure}

GCN in Fig.~\ref{fig:GCNImplementation} (see detail structure in Fig.~\ref{fig:GCNDetails1}) is to extract the spatial features of topological graphs by using the topologically-stable relationship information. Meanwhile, after convolutional graph operation, pixels feature belonging to the same class in semantic segmentation will be close to each other in the feature manifold (see Fig.~\ref{fig:GCNDetails}). 
		
We multiplied the feature map with the corresponding transpose as input of the GCN block.  
Global features will be generated by two layers of GCN blocks (see Fig.~\ref{fig:GCNDetails1}), while the global topological relationship of graph structure-based features (see Fig.~\ref{fig:GCNDetails}) is obtained. The final feature map is fused by adding (see Fig.~\ref{fig:GCNImplementation}) the encoder output and the global relationship node feature together.

\begin{figure}[t]
    \centering
    \includegraphics[width=1.0\textwidth]{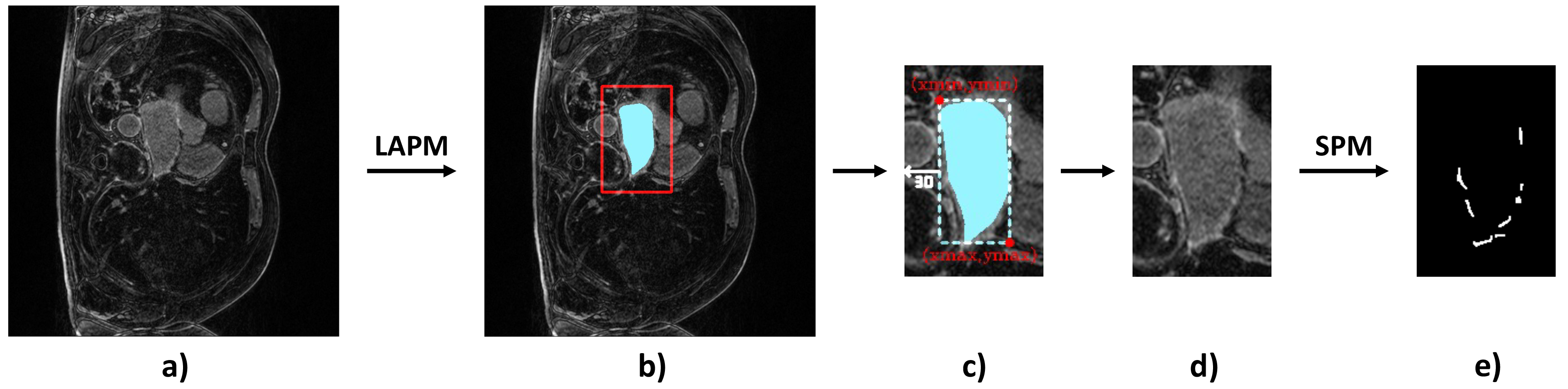}
    \caption{\textbf{Task 2} scar segmentation procedures: (a). LAMP Input, (b), Predicted LA, (c). Cropping positions, (d). Cropped ROI and SPM Input, and (e). Predicted Scar}
    \label{fig:task 1_pipeline}
\end{figure}

\section{Implementation}
\subsection{Dataset and Pre-processing}
The LAScarQS dataset includes two tasks: 1). LA and LA Scar segmentation (\textbf{task 1}), and 2). LA Segmentation across scanners (\textbf{task 2}). The first task contains 60 3D LGE-MRIs with labels containing LAs and LA scars, while the second consists of 130 3D LGE-MRIs from multiple medical centers with labels containing only LAs~\cite{fudan.edu.cn}.

In \textbf{task 1}, 54 subjects (approx. 44 slices per subject) are involved in the training test, while the remaining 6 subjects are used in the validation set. 
In \textbf{task 2}, 117 (approx. 44 slices per subject) and 13 subjects are used in the training and testing, respectively. 
Black margins are cropped, while images are resized to $224 \times 224$ with the bilinear interpolation before being normailzed to the range of [0, 1] by the min-max normalization.
Each image is augmented 4 times by random rotation with angles sampled from [$0\degree$, $180\degree$] and translation less than $0.1\cdot w$, where $w$ represents the image width.
The prediction performance is reported based on the 10 testing subjects available.

% Additionally, to amplify the dataset to improve robustness of the network, each training image is augmented to 4 through random affine transformation with rotation angle $[0, 180]$ and maximum absolute fraction of translations $0.1$.

\subsection{Training details}
We first trained the LA segmentation on \textbf{task 2}. The obtained model was loaded as the pre-training model for \textbf{task 1}. 
In detail, in the initial stage, the segmentation model was trained with all the LA labels available, obtaining the LA prediction model (LAPM). Then, we used the LAPM to roughly segment the targetted LA region, according to which images in the training set were cropped to train the scar prediction model (SPM). 
Specifically, the cropping region of interest (ROI) was implemented via $((x_{min}-30,y_{min}-30),(x_{max}+30,y_{max}+30))$, while $x_{min}$, $x_{max}$, $y_{min}$, $y_{max}$ were boundary pixels of the predicted LA region, $30$ was an empirically-selected tolerance of LA prediction. 
Finally, the prediction map was restored to its original size using zero padding.

% Our training procedure of Task 1 is organized into two stages to further utilize the relationship between the location of LA and scar, shown in \ref{task 1_pipeline}. We first train with only the LA labels and then crop all the datasets according to the LA predictions. Since the LA prediction is usually an irregular polygon and may not be accurate, which leads to further errors on the scar segmentation, we take the rectangle of$((x_{min}-30,y_{min}-30),(x_{max}+30,y_{max}+30))$ to crop the region of interest (ROI) to enhance the generalization capability.[30 for fault detection] The clipped slice and the label of the scar are then fed into the model, trained to produce the prediction of the scar. 

We implemented our network with the PyTorch library~\cite{NEURIPS2019_9015}. We ran 30 epochs on one NVIDIA Geforce RTX 3080Ti GPU. The batch size was 8, and the SGD optimizer was used. The initial learning rate was set as $10^{-4}$, which would be decayed to the previous 0.1 times when the validation dice records were updated. 
% When training the scar segmentation, we first train with only the LA labels and then crop all the datasets according to the LA predictions. The LA prediction is usually an irregular polygon and we take the rectangle of $((x_{min}-30,y_{min}-30),(x_{max}+30,y_{max}+30))$ as the crop region.[30 for fault detection] The clipped slice and the label of the scar are then fed into the model, trained to produce the prediction of the scar. Finally the prediction map will be restored to its original size.
% We ran 30 epochs on an NVIDIA 3080Ti GPU ...
% [test time for one epoch, pytorch framework, training lr, decay, epoch, batch size, training time... ]

\section{Experiment}
On both tasks, we compared our UGformer with other SOTA models, including U-Net~\cite{ronneberger2015u}, Res-U-Net~\cite{diakogiannis2020resunet}, Attention-U-Net~\cite{oktay2018attention}. 
%We also conducted ablation studies in scenarios including UGformer with / w.o. GCN. 
We also performed ablation studies to demonstrate the effectiveness of our EBT and GCN bridge modules. 
%The prediction performance is evaluated by Dice Scores (DS).
From obtained results demonstrated in Table~\ref{tab:tab1}, Table~\ref{tab:tab2}, and Table~\ref{tab:tab3}, we found that in both \textbf{task 1} and \textbf{task 2}, the proposed UGformer outperforms other baselines where transformers are engaged when evaluated by the Dice Score (DS).

\subsection{Comparison to the state-of-the-art methods (SOTA)}
\subsubsection{LA on Task 1 and Task 2}
\begin{figure}[!htb]
	\centering
	\subfigure[Image]{
		\includegraphics[width=0.3\textwidth]{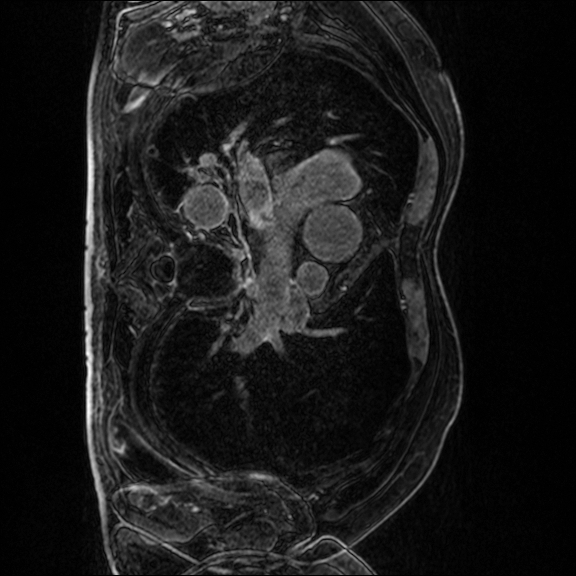}
		\label{subfig:Image la} }
	\subfigure[Gound Truth] {
		\includegraphics[width=0.3\textwidth]{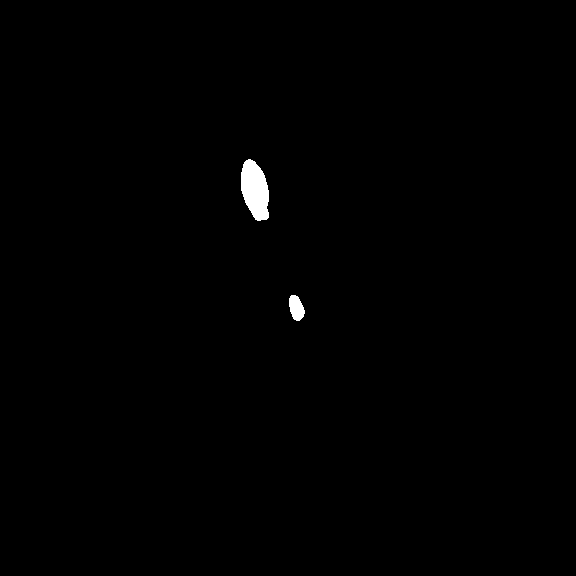}
		\label{subfig:Gound Truth la} }  
	\subfigure[U-Net] {
		\includegraphics[width=0.3\textwidth]{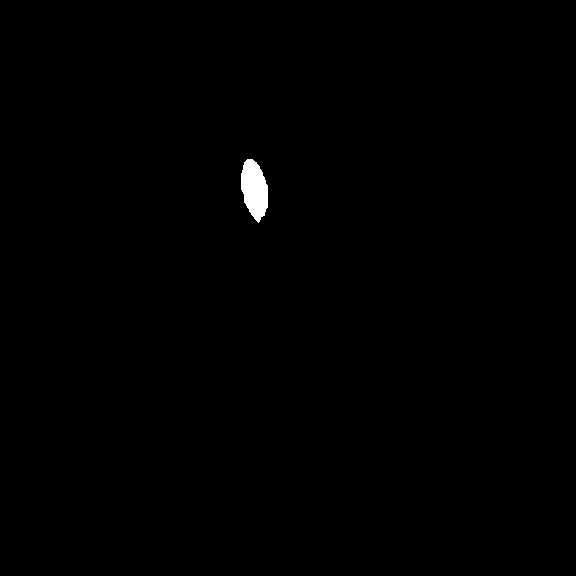}
		\label{subfig:U-Net la} }  
	\subfigure[Res-U-Net] {
		\includegraphics[width=0.3\textwidth]{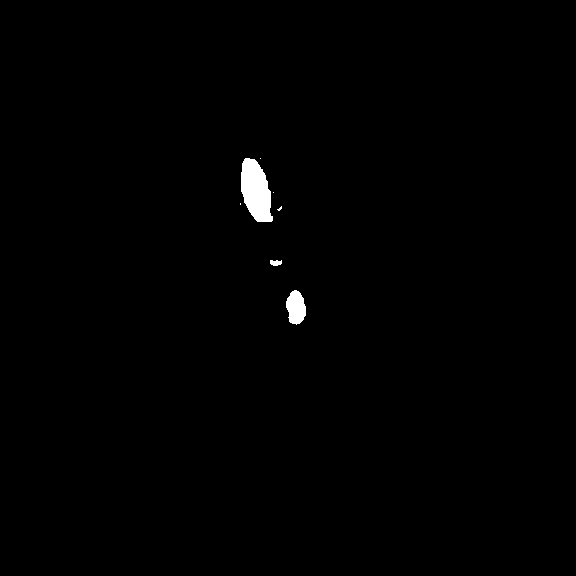}
		\label{subfig:Res-U-Net la } }  
	\subfigure[Attention-U-Net] {
		\includegraphics[width=0.3\textwidth]{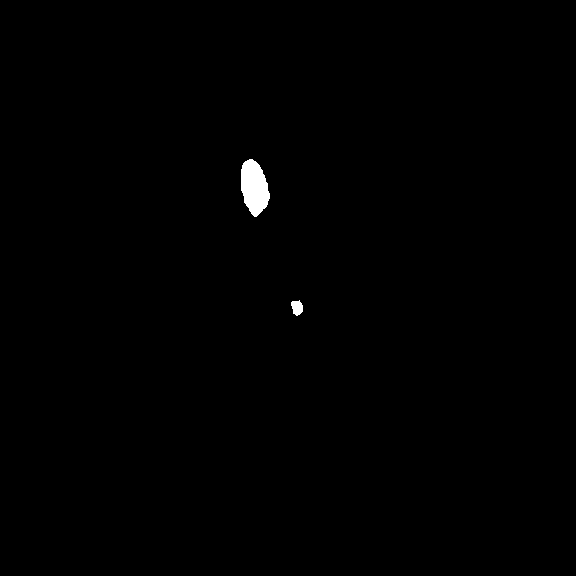}
		\label{subfig:Attention-U-Net la} }  
	\subfigure[UGformer] {
		\includegraphics[width=0.3\textwidth]{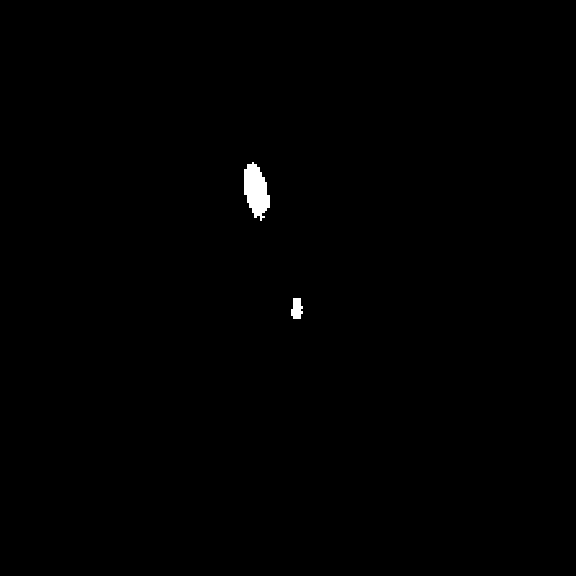}
		\label{subfig:UGformer la} } 
	\caption
	{Prediction results on task 2 LA.}
	\label{fig:la}
\end{figure}

:
% From Table~\ref{tab:tab1}, we can obverse that U-Net performs the best. UGformer is a close second and can outperform other SOTA models. At the same time, UGformer without the GCN module will work better than with this module, the reasons will be explained in section 4.2.
In Table~\ref{tab:tab1}, the dice scores outside before parentheses are performance by the model trained only with \textbf{task 1} LA dataset, while the numbers in brackets present results of models pre-trained by \textbf{task 2} dataset. 
We can clearly obverse that UGformer presents better prediction accuracy when predicting the LAs. Specifically, the proposed UGformer achieves the highest dice in \textbf{task 2}, outperforming all involved baselines. 
As shown in Figure~\ref{fig:la}, the proposed UGformer is capable of predicting small pathological areas. At the same time, unlike Res-U-Net, UGformer is able to avoid most false detection. We believe that such an appealing factor is brought by the fact that transformers are more sensitive to irregularly shaped pathological regions~\cite{xiao2021early}, while the GCN module further enhances the predictive power to small regions.

We can also find from Table~\ref{tab:tab1} that the Attention-U-Net performs the best no matter whether the pre-training stages are presented or not. 
In the meanwhile, if initialized by the pre-trained model, the DS of all the involved approaches is approx. 92 and 93. It is because that LA segmentation of \textbf{task 1} is a relatively simple assignment with consistent style information since they are generated from one single scanner.

\subsubsection{Scar on Task 1}
\begin{figure}[!htb]
	\centering
	\subfigure[Image]{
		\includegraphics[width=0.3\textwidth]{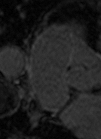}
		\label{subfig:Image sacr} }
	\subfigure[Gound Truth] {
		\includegraphics[width=0.3\textwidth]{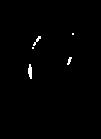}
		\label{subfig:Gound Truth scar} }  
	\subfigure[U-Net] {
		\includegraphics[width=0.3\textwidth]{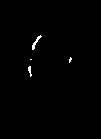}
		\label{subfig:U-Net scar} }  
	\subfigure[Res-U-Net] {
		\includegraphics[width=0.3\textwidth]{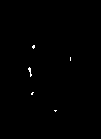}
		\label{subfig:Res-U-Net scar} }  
	\subfigure[Attention-U-Net] {
		\includegraphics[width=0.3\textwidth]{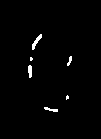}
		\label{subfig:Attention-U-Net scar} }  
	\subfigure[UGformer] {
		\includegraphics[width=0.3\textwidth]{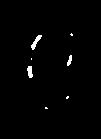}
		\label{subfig:UGformer scar} } 
	\caption
	{Prediction results on task 1 Scar. Res-U-Net can not predict the pathology. U-Net and Attention-U-Net can predict a certain part of the pathology. Nevertheless, we can also obverse worse false detection than that predicted by the proposed UGformer.}
	\label{fig:scarimg}
\end{figure}

:
The proposed UGformer performs the best in this scenario by at least 2.5\% compared to other baselines. It demonstrates that it is particularly useful in quantifying irregular and scattered LA scars. 
% Its dice score exceeds that of the best baseline model by 2.5 percent. 
As shown in Fig.~\ref{fig:scarimg}, UGformer clearly identifies more pathological regions and contributes to fewer false detections. 
% Attention-U-Net also predicts a certain part of pathology, nevertheless, we can also obverse worse false detection than that predicted by the proposed UGformer.

% lower left corner is significantly wider than the label in ground truth, resulting in false detections.

\begin{table}[]\centering
\caption{Comparison between SOTA models.}\label{tab:tab1}
\begin{tabular}{|c|c|c|c|}
\hline
\multirow{2}{*}{Method} & \textbf{Task 1-LA} & \textbf{Task 1-Scar} & \textbf{Task 2-LA} \\ \cline{2-4} 
                        & DS$\uparrow$     & DS$\uparrow$       & DS$\uparrow$     \\ \hline
U-Net                   & 85.95 (92.24)    & 67.76      & 84.42   \\ \hline
Res-U-Net               & 85.26 (92.28)    & 62.61      & 83.74    \\ \hline
Attention-U-Net         & 87.40 (\textbf{93.22)}   & 70.11      & 85.37    \\ \hline
%UGformer w.o. GCN       & 84.47(92.33)    & 70.82      & 85.44    \\ \hline
UGformer                & 85.49 (92.36)  & \textbf{72.66}     & \textbf{86.59}    \\ \hline
\end{tabular}
\end{table}

% \begin{figure}[!htb]
% 	\centering
% 	\subfigure[Image]{
% 		\includegraphics[width=0.4\textwidth]{predict_scar/slice22_enhanced.png}
% 		\label{subfig:Image sacr} }
% 	\subfigure[Gound Truth] {
% 		\includegraphics[width=0.4\textwidth]{predict_scar/slice22_scar.png}
% 		\label{subfig:Gound Truth scar} }  
% 	\subfigure[U-Net] {
% 		\includegraphics[width=0.4\textwidth]{predict_scar/unetscar.png}
% 		\label{subfig:U-Net scar} }  
% 	\subfigure[Res-U-Net] {
% 		\includegraphics[width=0.4\textwidth]{predict_scar/resscar.png}
% 		\label{subfig:Res-U-Net scar} }  
% 	\subfigure[Attention-U-Net] {
% 		\includegraphics[width=0.4\textwidth]{predict_scar/attuscar.png}
% 		\label{subfig:Attention-U-Net scar} }  
% 	\subfigure[UGformer] {
% 		\includegraphics[width=0.4\textwidth]{predict_scar/finalscar.png}
% 		\label{subfig:UGformer scar} } 
% 	\caption
% 	{Prediction results on task 1 Scar. U-Net and Res-U-Net almost can not predict the pathology. Attention-U-Net can predict a certain part of the pathology. Nevertheless, we can also obverse worse false detection than that predicted by the proposed UGformer.}
% 	\label{fig:scarimg}
% \end{figure}

\subsection{Ablation studies}
% \subsubsection{Influence of U-shape}
% we first ablate the impacts of the U-shaped architecture. Using the convolutions as the decoders significantly enhances the performance. The results are shown in Table 1. Compared to the model without U-shape design, adding convolutional decoders improve the dice score on the validation set by [], decrease the Hausdorff Distance and Average Surface Distance  score by [].

% \begin{table}[]
% \caption{Impacts of U-shaped architecture(skip connection).}\label{tab1}
% \begin{tabular}{|c|ccc|cll|}
% \hline
% \multirow{2}{*}{Model} & \multicolumn{3}{c|}{Task1} & \multicolumn{3}{c|}{Task2} \\ \cline{2-7} 
%                   & \multicolumn{1}{c|}{Dice$\uparrow$} & \multicolumn{1}{c|}{HD$\downarrow$} & ASD$\downarrow$ & \multicolumn{1}{c|}{Dice$\uparrow$} & \multicolumn{1}{l|}{HD$\downarrow$} & ASD$\downarrow$ \\ \hline
% U-shaped UGformer & \multicolumn{1}{c|}{}     & \multicolumn{1}{c|}{}   &     & \multicolumn{1}{c|}{}     & \multicolumn{1}{l|}{}   &     \\ \hline
% UGformer          & \multicolumn{1}{c|}{}     & \multicolumn{1}{c|}{}   &     & \multicolumn{1}{c|}{}     & \multicolumn{1}{l|}{}   &     \\ \hline
% \end{tabular}
% \end{table}

\subsubsection{Influence of ETB module}
In Table~\ref{tab:tab2}, ablations of MHSAs and DCs in the ETB are presented.
We can conclude that both MHSAs and DCs are essential to achieve the best segmentation performance at 85.49\%, 72.66\%, and 86.59\% on DS on \textbf{task 1-LA}, \textbf{task 1-Scar}, and \textbf{task 2-LA}, respectively. 
Particularly, the combination of MHSAs and DCs module makes the greatest significant improvement on \textbf{task 2-LA} by 7\%. It proves that the two modules contribute to each other and help the prediction of the model.

% it can be concluded that the combining the two attention modules is the most suitable method for the ETB. IT has the highest dice and the lowest HD score. for LA prediction in both task 1 and task 2. It also has the highest dice in the scar of task 1.

\begin{table}[]
\caption{Comparison of ETB module.}
\label{tab:tab2}\centering

% \centering
\begin{tabular}{|c|c|c|c|c|}
\hline
\multirow{2}{*}{MHSA} & \multirow{2}{*}{DC} & \textbf{Task 1-LA} & \textbf{Task 1-Scar} & \textbf{Task 2-LA} \\ \cline{3-5} 
                      &                     & DS       & DS         & DS     \\ \hline
\checkmark            &                     & 85.06    & 69.65      & 78.67    \\ \hline
                      & \checkmark          & 85.26    & 70.50      & 80.66    \\ \hline
\checkmark            & \checkmark          & \textbf{85.49}  & \textbf{72.66}     & \textbf{86.59}    \\ \hline
\end{tabular}
\end{table}

\subsubsection{Influence of GCN}
Table~\ref{tab:tab3} enumerates the results of ablations of GCN block when the proposed UGformer and U-Net are used as backbones. 
From there, we can find that GCN improves the prediction performance of U-Net in \textbf{task 1-LA} and \textbf{task 2-LA}. 
However, the improvement in scar prediction in \textbf{task 1-Scar} with U-Net is insignificant.
When GCN is implemented in the UGformer architecture, it improves the prediction performance in all settings. Particularly, when predicting scars, GCN module improves the transformer performance from 70.82\% to 72.66\% by 2.6\%.

% Since the U-Net is a simple CNN model, GCN bridge would help it to model the global relationship and achieve better segmentation results than baseline. However, for UGformer, which itself consists of attention modules, adding a GCN bridge may not have such a big upgrade. 

\begin{table}[!htbb]
\caption{Comparison of different bridge module.}\centering\label{tab:tab3}

\begin{tabular}{|c|c|c|c|c|}
\hline
\multirow{2}{*}{Architecture} & \multirow{2}{*}{GCN} & Task 1-LA & Task 1-Scar & Task 2-LA \\ \cline{3-5} 
                              &                      & DS       & DS         & DS       \\ \hline
\multirow{2}{*}{U-Net}        &                      & 85.95    & \textbf{67.76}      & 84.42    \\ \cline{2-5} 
                              & \checkmark           & \textbf{87.93}    & 67.72      & \textbf{86.79}    \\ \hline
\multirow{2}{*}{UGformer}     &                      & 84.47   & 70.82      & 85.44    \\ \cline{2-5} 
                              & \checkmark           & \textbf{85.49}  & \textbf{72.66}     & \textbf{86.59}    \\ \hline
\end{tabular}
\end{table}

%From the data, we can find that GCN improves prediction performance of Unet in all the tasks and improves the prediction performance of UGformer in only task 1 - LA. Since the Unet is a simple model consisting of common convolutions, the addition of the GCN bridge is equivalent to adding an attention module to the unet, which helps it to obtain more detailed information and improve its performance. However, for UGformer, which itself consists of attention modules, adding a GCN bridge to a data set with multiple scanners like task 2-LA and a small target like task 1-Scar may affect the extraction of detailed information. However, for task 1-LA, where the data is simple, it can help to train the model to a certain extent.

\subsubsection{Influence of the two-stage method}
\begin{figure}[!htbb]
	\centering
	\subfigure[Original Image]{
		\includegraphics[width=0.3\textwidth]{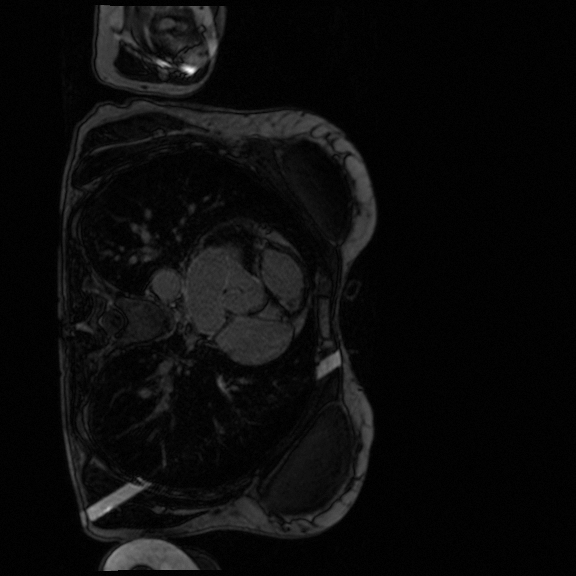}
		\label{subfig:Image} }
	\subfigure[Original Gound Truth] {
		\includegraphics[width=0.3\textwidth]{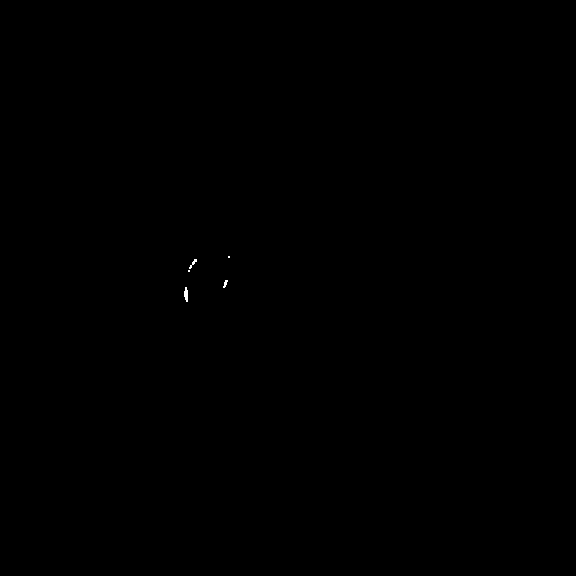}
		\label{subfig:Gound Truth} }  
	\subfigure[Normal Preditcion] {
		\includegraphics[width=0.3\textwidth]{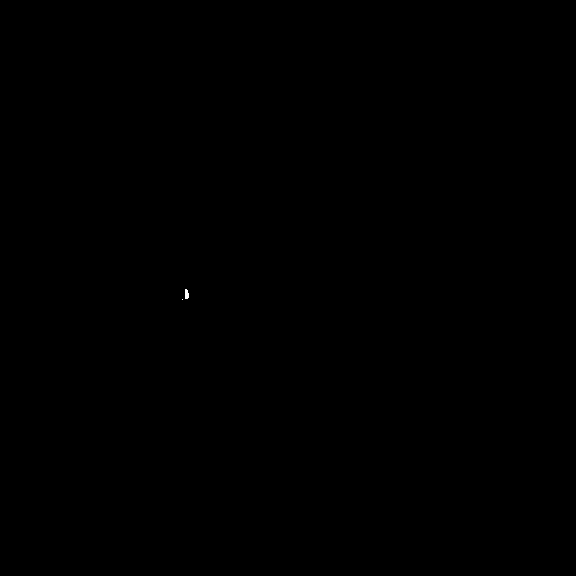}
		\label{subfig:U-Net-normal} }  
% 	\\ \hline
	\subfigure[Cropped Image] {
		\includegraphics[width=0.3\textwidth]{twostage/crop_enhance.png}
		\label{subfig:Res-U-Net} }  
	\subfigure[Cropped Ground Truth] {
		\includegraphics[width=0.3\textwidth]{twostage/crop_scar.png}
		\label{subfig:Attention-U-Net} }  
	\subfigure[Two Stage Prediction] {
		\includegraphics[width=0.3\textwidth]{twostage/ugscar.png}
		\label{subfig:UGformer-twosatge} } 
	\caption
	{Prediction results on original images and cropped images}
	\label{fig:twostage}
\end{figure}

Figure~\ref{fig:twostage} displays the prediction results with the two-stage prediction approaches and the normal ones. It can be clearly seen that the two stage method has successfully predicted most of the scars (see Fig.~\ref{subfig:U-Net-normal}), although some kind of false detection can still be observed. Nevertheless, with the common prediction method (see Fig.~\ref{subfig:UGformer-twosatge}), the scar is almost impossible to be predicted. 
We can hereby conclude that the two-stage prediction approach is essential in quantifying scars with irregular and tiny occupations on the picture.

\section{Conclusions}
In this paper, we proposed the UGformer, a novel U-shaped transformer architecture with a GCN bridge. It is capable of segmenting the left atrium (LA) across different scanners and quantifying LA scars with a two-stage predicting strategy given late gadolinium enhancement magnetic resonance images. 
Specifically, an enhanced transformer block combining multi-head self-attention and deformable convolutions is introduced to model global attention and overcome degradation in quantifying scars with irregular shapes. 
We also employ a graph convolution network (GCN), a novel GCN-based bridge, to optimize the global space of intermediate feature layers. 
Extensive empirical experiments on the LAScarQS 2022 challenge dataset have demonstrated the effectiveness and robustness of the proposed UGformer architecture in LA prediction and scar quantification.

\bibliographystyle{splncs04}
\bibliography{reference.bib}

\end{document}